\documentclass[acmtog]{acmart} 

\acmSubmissionID{431}

\usepackage{booktabs}
\usepackage{graphicx}
\usepackage{enumitem}
\usepackage{soul}
\usepackage{dsfont}
\usepackage{gensymb}
\usepackage{microtype}
\usepackage{pifont}
\usepackage{threeparttable}

\usepackage{amssymb}
\usepackage{multirow}
	
\usepackage{color, colortbl}

\newcommand{\ie}{i.e.,\ }

\definecolor{aliceblue}{rgb}{0.94, 0.97, 1.0}

\newcolumntype{a}{>{\columncolor{aliceblue}}r}

\acmJournal{TOG}

\newcommand{\rev}[1]{\textcolor{black}{#1}}

\begin{document}

\setcopyright{acmlicensed}
\acmJournal{TOG}
\acmYear{2023} \acmVolume{42} \acmNumber{6} \acmArticle{} \acmMonth{12} \acmPrice{}\acmDOI{10.1145/3618357}

\title{Diffusion Posterior Illumination for Ambiguity-aware Inverse Rendering}
\author{Linjie Lyu}
\email{llyu@mpi-inf.mpg.de}
\orcid{0009-0007-4763-8457}
\affiliation{%
  \institution{Max-Planck-Institut für Informatik}
  \country{Germany}
}
\author{Ayush Tewari}
\email{ayusht@mit.edu}
\affiliation{%
  \institution{MIT CSAIL}
  \country{USA}
}
\author{Marc Habermann}
\email{mhaberma@mpi-inf.mpg.de}
\affiliation{%
  \institution{Max-Planck-Institut für Informatik}
  \country{Germany}
}
\author{Shunsuke Saito}
\email{shunsuke.saito16@gmail.com}
\affiliation{%
  \institution{Reality Labs Research}
  \country{USA}
}
\author{Michael Zollhöfer}
\email{zollhoefer@meta.com}
\affiliation{%
  \institution{Reality Labs Research}
  \country{USA}
}
\author{Thomas Leimkühler}
\email{thomas.leimkuehler@mpi-inf.mpg.de}
\affiliation{%
  \institution{Max-Planck-Institut für Informatik}
  \country{Germany}
}
\author{Christian Theobalt}
\email{theobalt@mpi-inf.mpg.de}
\affiliation{%
  \institution{Max-Planck-Institut für Informatik}
  \country{Germany}
}
\begin{abstract}
Inverse rendering, the process of inferring scene properties from images, is a challenging inverse problem.
The task is ill-posed, as many different scene configurations can give rise to the same image. 
Most existing solutions incorporate priors into the inverse-rendering pipeline to encourage plausible solutions, but they do not consider the inherent ambiguities and the multi-modal distribution of possible decompositions.
In this work, we propose a novel scheme that integrates a denoising diffusion probabilistic model pre-trained on natural illumination maps into an optimization framework involving a differentiable path tracer. 
The proposed method allows sampling from combinations of illumination and spatially-varying surface materials that are, both, natural and explain the image observations. 
We further conduct an extensive comparative study of different priors on illumination used in previous work on inverse rendering.
Our method excels in recovering materials and producing highly realistic and diverse environment map samples that faithfully explain the illumination of the input images.
\end{abstract}
%


\begin{teaserfigure}
	\centering
	\includegraphics[width=\linewidth]{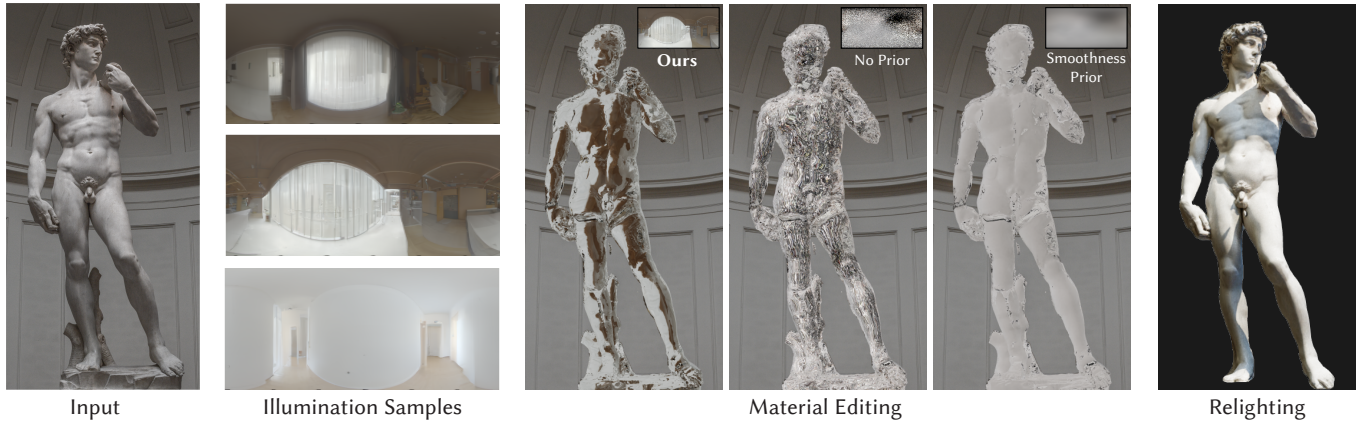} 
	\caption
	{
    Given one or multiple images of a scene (\emph{Input}), our method infers surface materials and illumination.
    We model the inherent ambiguities involved in this process:
    The statue could have been lit by an infinite number of environment maps, all giving rise to the same image.
    Our diffusion prior allows to sample from a joint distribution of materials and natural illumination conditions (\emph{Illumination Samples}) that all explain the input.
    This enables high-quality \emph{Material Editing} (here demonstrating a change to a translucent material; environment map reconstructions are shown as insets) and \emph{Relighting}.
    Without any prior, the recovered illumination is noisy, the commonly used smoothness prior converges to a single blurry result.
    In contrast, our samples are diverse and natural.
	}
	\label{fig:teaser}
\end{teaserfigure}

\maketitle

%
%
\section{Introduction}
\label{sec:introduction}
%

Inverse rendering is the process of inferring scene properties such as geometry, lighting, and surface materials from images.
It is a long-standing challenge with applications in scene understanding, image editing, urban planning, virtual reality, and many more.
Typically, inverse-rendering pipelines employ an analysis-by-synthesis approach, in which scene parameters are optimized via a differentiable renderer based on an image reconstruction objective.
Unfortunately, determining scene properties from (multi-view) images alone is a severely ill-posed inverse problem since light transport is governed by the rendering equation \cite{kajiya1986rendering}, which connects geometry, materials, and illumination via complex integral relationships.
This naturally leads to ambiguities:
A red pixel might arise from a red material under white illumination, or from a white material under red illumination;
a smooth-looking surface might arise from a diffuse material or low-frequency illumination, etc.

Fortunately, real scenes exhibit statistical regularities \cite{dror2004statistical}.
Exploiting the fact that some scene configurations are more likely than others, most existing solutions incorporate \emph{priors} into the inverse-rendering pipeline.
They are meant to encourage plausible solutions. 
When considering the joint distribution of illumination and materials, applying a prior over illumination conditions naturally narrows down the distribution over materials, and vice versa.
Priors vary in their degree of sophistication, ranging from simple heuristics such as smoothness \cite{zhang2021nerfactor} to the use of data-driven deep generative models \cite{gardner2022rotation}.
However, the inherent multi-solution ambiguities of the inverse-rendering problem are usually not considered:
With a naive prior, the optimization tends to converge to \emph{one} local optimum, which is determined by the weight given to the prior, effectively balancing the trade-off between re-rendering accuracy and the naturalness of the obtained solution \cite{blau2018perception}.
Ideally, an inverse-rendering prior should not only \emph{faithfully capture} the distribution of scene properties, such as illumination or material, but also enable the \emph{the generation of diverse solutions from the multi-modal posterior distribution} of scene properties given the input images. Typically, one must choose between achieving high quality or diversity, but a well-designed prior can achieve both simultaneously.

Diffusion models \cite{sohl2015deep} excel at capturing data distributions and have demonstrated the ability to generate realistic and diverse samples. A trained unconditional diffusion model can be utilized to solve non-linear inverse problems \cite{chung2022diffusion}.
Therefore, in this paper, we investigate the use of diffusion posterior sampling (DPS) as a prior for inverse rendering. 
To this end, we train a denoising diffusion probabilistic model (DDPM) \cite{ho2020denoising} on natural environment maps and then integrate it into an optimization framework involving a differentiable path tracer.
Crucially, we extend DPS to incorporate a measurement function with trainable parameters.
Our novel optimization scheme allows sampling from combinations of illumination and spatially-varying materials that are \emph{natural}, \emph{diverse}, and \emph{explain the image observations}. 

We further conduct a thorough investigation into the effectiveness of various illumination priors for inverse rendering.
We demonstrate that our novel formulation not only surpasses the state of the art in its ability to produce highly realistic and diverse environment map samples while fitting the input data in the form of monocular and multi-view images.
It also simultaneously facilitates disentanglement and reduces ambiguities in the optimized materials. 

In summary, our contributions are:
\begin{itemize}
\item{A method for recovering the posterior distribution of illumination and materials from image observations under an unknown lighting condition, hence generating samples with both high realism and high diversity.}
\item{A novel strategy combining the denoising process in DDPM and differentiable rendering for the joint optimization of scene illumination and materials.}
\item{A comprehensive study of different priors on illumination used in inverse rendering.}
\end{itemize}

\section{Related Work}
\label{sec:relatedwork}
%
%

We review closely related literature on differentiable rendering methods (Sec.~\ref{sec:previous_rendering}) and popular priors on illumination that are widely used in inverse rendering (Sec.~\ref{sec:previous_priors}). Then, we briefly discuss the usage of diffusion generative models for general inverse problems (Sec.~\ref{sec:previous_diffusion}).


\subsection{Differentiable and Inverse Rendering}
\label{sec:previous_rendering}

Inverse rendering is commonly performed by means of an analysis-by-synthesis approach: 
A forward rendering model turns scene parameters such as geometry, illumination, and surface materials into an image, which is compared to existing image observations.
For gradient-based optimization of scene parameters to work, the rendering model needs to be (made) differentiable.

Approximate differentiable rendering approaches that rely on mesh rasterization have been proposed for joint optimization of material and illumination.
By design, they either ignore global light transport \cite{kato2018neural,liu2019soft,loper2014opendr,laine2020modular} or approximate the rendering function of indirect illumination to account for isolated effects like soft shadows \cite{lyu2021efficient}.


In contrast, more physically-based approaches \cite{li2018differentiable,nimier2019mitsuba,loubet2019reparameterizing, Jakob2020DrJit} account for global illumination by differentiating Monte Carlo-sampled multi-bounce light transport.
Special precautions are required to obtain unbiased gradients arising from non-differentiable visibility \cite{li2018differentiable,loubet2019reparameterizing}. A physically-based differentiable path tracer is powerful in inverse rendering with complex global light transport effects \cite{hasselgren2022shape}, but can suffer from high-frequency noise in the gradients.

Recent years have witnessed a shift towards neural implicit representations in differentiable and inverse rendering.
Continuous volume densities \cite{mildenhall2021nerf} and signed distance functions (SDFs) \cite{park2019deepsdf} have many advantages over classical representations, in particular when it comes to differentiability.
However, the resulting improved disentanglement of geometry, material, and illumination in the neural rendering process comes at the cost of increased computational complexity.
This has spurred research interest in hybrid representations \cite{Munkberg_2022_CVPR}.
In particular, the incorporation of global illumination into the (inverse) rendering process is costly and demands approximations \cite{nerv2021}.


In this work, we build upon the state-of-the-art differentiable path tracer Mitsuba 3 \cite{jakob2022mitsuba3,Jakob2020DrJit}, which allows for efficient differentiation \rev{of} its path-tracing process. 
Providing a principled and scalable solution for modeling global illumination, we combine it with a strong diffusion-based illumination prior to recover a distribution over spatially-varying materials and illumination. 

Compact low-frequency illumination models based on spherical harmonics (SH) \cite{basri2003lambertian,ramamoorthi2001efficient} are popular in traditional real-time rasterization methods. 
Spherical Gaussians \cite{wang2009all} excel in representing a sparse set of high-frequency features and have recently been used for neural inverse rendering \cite{zhang2021physg,zhang2022modeling,wu2023nefii,jin2023tensoir}.
However, this representation does not scale favorably to complex natural illumination conditions.
In this work, we focus on a regularly sampled 2D environment map under equirectangular projection, which matches well with current CNN-based diffusion models.






\subsection{Priors for Inverse Rendering}
\label{sec:previous_priors}

Inverse-rendering problems are notoriously ill-posed, and various priors have been proposed to address this issue. Data-\emph{agnostic} priors include smoothness regularization, as utilized by NeRFactor ~\cite{zhang2021nerfactor} and NRTF ~\cite{lyu2022neural}, which mitigates high-frequency noise in estimated environment maps. NvDiffrecMC~\cite{hasselgren2022shape} proposes a prior that assumes illumination is mostly monochrome, promoting high-frequency lighting and sharp shadows represented by lighting instead of material textures. \rev{The Deep Image Prior} (DIP)~\cite{ulyanov2018deep} demonstrates the effectiveness of randomly initialized neural networks as handcrafted priors, achieving excellent results in standard inverse problems such as denoising, super-resolution, and inpainting. Data-agnostic priors are effective heuristics, but they may not accurately reflect the true underlying distribution of the data. The performance of these priors is highly dependent on the choice of hyper-parameters, which can be difficult to tune. Additionally, using these priors often leads to optimization algorithms converging to a single local minimum, resulting in a loss of diversity in the final solutions.

On the other hand, data-\emph{driven} priors have the advantage of more accurately capturing the true distribution of real illumination environments. They have the potential to enable sampling from the multi-modal posterior distribution, providing a more diverse set of solutions. \citet{egger2018occlusion} as well as \citet{yu2021outdoor} utilize a linear statistical model with a Gaussian prior in the space of spherical harmonics (SH) coefficients to avoid unrealistic illumination environments. However, this approach can only reproduce low-frequency lighting effects. Emlight~\cite{zhan2021emlight} decomposes the illumination map into spherical light distribution, light intensity, and the ambient term for natural illumination regression. Stylelight~\cite{wang2022stylelight} and ImmerseGAN~\cite{dastjerdi2022guided} propose GAN panorama generation networks for realistic lighting estimation and editing. \citet{gardner2022rotation} develop a rotation-equivariant, high-dynamic-range (HDR) neural illumination model based on a variational auto-decoder (VAE)~\cite{rezende2015variational} that can express complex features of the natural environment distribution and use it for inverse rendering. VAEs and GANs are powerful techniques for generating diverse samples of illumination that can be used to solve inverse problems, given input images. However, while these methods can provide a range of solutions, the quality of the rendered images and the naturalness of the sampled environment maps \rev{are inferior to what our approach delivers}.


\subsection{Generative Diffusion Models for Inverse Problems}
\label{sec:previous_diffusion}

Denoising diffusion probabilistic models (DDPMs)~\cite{sohl2015deep,ho2020denoising,song2020score} have recently demonstrated remarkable performance on both faithfulness and diversity of samples in tasks like image synthesis~\cite{dhariwal2021diffusion}. 
Also, a pre-trained, task-agnostic DDPM has been demonstrated to be an excellent prior for imaging inverse problems:
Typically, one can resort to iterative projections to the measurement space~\cite{choi2021ilvr,chung2022come,song2020score} or estimate the posterior score function \cite{chung2022diffusion} to reach feasible solutions from the implicit prior data distribution.
This has been used both in linear \cite{kawar2022denoising,chung2022come,song2020score} and non-linear \cite{chung2022diffusion,song2023pseudoinverse} imaging inverse problems. 
While most of the inverse problem solvers with DDPMs are limited to a known and fixed forward measurement operator, BlindDPS~\cite{chung2022parallel} proposes to sample the image and the operator parameters from posterior score functions in parallel, to solve blind inverse problems with unknown forward operators.
In contrast to our solution, they require an additional DDPM for the operator.

This work sets out to utilize DDPMs as a prior for natural illumination distribution to address complex non-linear inverse rendering problems. Specifically, we aim to sample diverse but realistic combinations of illumination and materials that can accurately explain the given observation, which poses significant challenges.
%
%
\section{Background}
\label{sec:background}
In this section, we recap Denoising Diffusion Probabilistic Models (DDPMs) (Sec.~\ref{sec:ddpm}), followed by an introduction of the diffusion posterior sampling method (Sec.~\ref{sec:dps}) for general inverse problems. 

\subsection{Denoising Diffusion Probabilistic Models}
\label{sec:ddpm}

Denoising diffusion probabilistic models (DDPMs) \cite{ho2020denoising,sohl2015deep} are a type of generative models that define the generation process as a Markov chain of denoising steps. Formally, given the clean data $\mathbf{x}_0 \sim q\left(\mathbf{x}_0\right)$, in the forward steps, DDPMs gradually add noise to the data, which is described as a Gaussian transition: 
%
%
\begin{equation}
\label{eq:ddpm_forward}
q\left(\mathbf{x}_t \mid \mathbf{x}_{t-1}\right):=\mathcal{N}\left(\mathbf{x}_t ; \sqrt{1-\beta_t} \mathbf{x}_{t-1}, \beta_t \mathbf{I}\right) \,,
\end{equation}  
%
%
where $0<\beta_1 <\beta_2<....< \beta_T=1$ are the fixed variance schedule. 
The distribution of sampling $x_t$ at the time step $t$ has a closed form 
%
%
\begin{equation}
\label{eq:ddpm_x_t}
q\left(\mathbf{x}_t \mid \mathbf{x}_0\right)=\mathcal{N}\left(\mathbf{x}_t ; \sqrt{\bar{\alpha}_t} \mathbf{x}_0,\left(1-\bar{\alpha}_t\right) \mathbf{I}\right) \,,
\end{equation}
%
%
with $\alpha_t=1-\beta_t \text { and } \bar{\alpha}_t=\prod_{s=1}^t \alpha_s $.
This enables efficient training since we can randomly sample a single timestep during training and generate the input $\mathbf{x}_t$. 
At the largest time step $T$, the noised data is transformed to a standard Gaussian distribution.

DDPMs learn the reverse Gaussian transition that gradually denoises the noised data with the objective:
%
%
\begin{equation}
\label{eq:ddpm_loss}
\mathcal{L}(\theta):=\mathbb{E}_{t, \mathbf{x}_0, \boldsymbol{\epsilon}}\left[\left\|\boldsymbol{\epsilon}-\boldsymbol{\epsilon}_\theta\left(\mathbf{x}_t, t\right)\right\|^2\right],
\end{equation}
%
%
corresponding to the ``simple'' loss formulation in \citet{ho2020denoising},
where $\epsilon_\theta$ denotes the residual noise predicted by a network and $\theta$ are the free variables of the network. 
The data at the next timestep can now be computed as:
%
%
\begin{equation}
\label{eq:ddpm_reverse}
\mathbf{x}_{t-1}=\frac{1}{\sqrt{\alpha_t}}\left(\mathbf{x}_t-\frac{1-\alpha_t}{\sqrt{1-\bar{\alpha}_t}} \boldsymbol{\epsilon}_\theta\left(\mathbf{x}_t, t\right)\right)+\sigma_t \mathbf{z} \,,
\end{equation}
%
%
where $\mathbf{z} \sim \mathcal{N}(\mathbf{0}, \mathbf{I})$ and the covariance $\sigma_t$ is fixed or learned. 
At test time, a randomly sampled distribution can be gradually denoised into the learned data manifold. 

\citet{song2020score} show that there is a continuous \rev{stochastic differential equation} (SDE)~\cite{anderson1982reverse} \rev{formulation} of the generation process,  which is equivalent to Eq.~\ref{eq:ddpm_reverse}:
%
%
\begin{equation}
\label{eq:sde_reverse}
d \boldsymbol{x}=\left[-\frac{\beta(t)}{2} \boldsymbol{x}-\beta(t) \nabla_{\boldsymbol{x}_t} \log p_t\left(\boldsymbol{x}_t\right)\right] d t+\sqrt{\beta(t)} d \overline{\boldsymbol{w}} \, .
\end{equation}
%
%
\rev{Thus, sampling from a DDPM can also be seen as numerically 
 solving the SDE.} Here, $d \overline{\boldsymbol{w}}$ is the standard Wiener process running backward.
The score function $\nabla_{\mathbf{x}_t} \log p_t\left(\mathbf{x}_t\right)$ can be approximated by a network $\boldsymbol{s}_\theta\left(\mathbf{x}_t,t\right)$ using score matching~\cite{hyvarinen2005estimation,vincent2011connection}, which is connected with the noise predictor in DDPMs via 
%
%
\begin{equation}
\label{eq:score_function}
\boldsymbol{s}_\theta\left(\mathbf{x}_t,t\right)=\frac{\boldsymbol{\epsilon}_\theta\left(\mathbf{x}_t, t\right)}{\sqrt{1-\bar{\alpha}_t}} \, .
\end{equation}
One can also find the posterior estimation of the clean image $\hat{\mathbf{x}}_t$ at the t step by
%
%
\begin{equation}
\label{eq:posterior_mean}
\hat{\mathbf{x}}_t=\dfrac{1}{\sqrt{\bar{\alpha}_t}}(\mathbf{x}_t+(1-\bar{\alpha}_t)\boldsymbol{s}_{\theta}(\mathbf{x}_t,t)) \,.
\end{equation}
%
%
%
For a comprehensive overview, we refer to \citet{mcallester2023mathematics}.
%
%
%
\subsection{Diffusion Posterior Sampling}
\label{sec:dps}
\citet{chung2022diffusion} proposed Diffusion Posterior Sampling (DPS)  to solve general inverse problems with a pre-trained unconditional DDPM. Consider the forward process
%
%
\begin{equation}
\label{eq:dps_measurement}
\mathbf{y}=\mathcal{M}(\mathbf{x}_0)+\mathbf{n},\quad\mathbf{y},\mathbf{n}\in\mathbb{R}^n,\:\mathbf{x}\in\mathbb{R}^d \,,
\end{equation}
%
%
where $\mathcal{M}(\cdot):\mathbb{R}^d\mapsto\mathbb{R}^n$ is the measurement operator, $\mathbf{x}_0$ is the input data, $\mathbf{y}$ is the measurement observation and $\mathbf{n}$ is measurement noise. In an inverse problem, we attempt to estimate the input image $\mathbf{x}_0$ given the observation $\mathbf{y}$ from the posterior distribution $p(\mathbf{x}_0 \mid \mathbf{y})$. 

\citet{chung2022diffusion} approximate the  posterior score function given observation $\mathbf{y}$ as:
%
%
\begin{equation}
\label{eq:posterior_score_function}
\nabla_{\mathbf{x}_t}\log p_t(\mathbf{x}_t \mid \mathbf{y})\simeq \boldsymbol{s}_{\theta}\left(\mathbf{x}_t,t\right)-\frac{1}{\sigma^{2}}\nabla_{\mathbf{x}_t}\|\mathbf{y}-\mathcal{M}(\hat{\mathbf{x}}_t)\|_2^2 \,.
\end{equation}
They assume a known and fixed forward operator $\mathcal{M}(\cdot)$ and Gaussian measurement noise $\mathbf{n} \sim \mathcal{N}(\mathbf{0}, \sigma^{2}\mathbf{I})$.
This posterior score function is then used to denoise the image in the reverse generation process: 
%
%
\begin{equation}
\label{eq:post_sde_reverse}
d \boldsymbol{x}=\left[-\frac{\beta(t)}{2} \boldsymbol{x}-\beta(t) \nabla_{\boldsymbol{x}_t} \log p_t\left(\boldsymbol{x}_t \mid \mathbf{y}\right)\right] d t+\sqrt{\beta(t)} d \overline{\boldsymbol{w}} \, .
\end{equation}
%
%

The posterior score function in Eq.~\ref{eq:posterior_score_function} aims to generate natural outputs while guiding them to fit the observation at the same time. 
However, DPS requires the operator $\mathcal{M}(\cdot)$ to be known and fixed, which is not the case for our method, where we seek to jointly optimize illumination and material parameters from images.
Here, the operator is the rendering function, \rev{which naturally encompasses both illumination and materials.}
We will present a method that enables jointly optimizing for the output of the diffusion model \rev{\ie scene illumination,} as well as the parameters of the measurement operator \rev{\ie scene materials}. 

%
%
%
\section{Method}
\label{sec:method}
Our method takes monocular or multi-view images of an object with a well-initialized geometry under one unknown illumination as input and allows realistic environment map sampling and joint optimization of material that fits the observation. To this end, our method not only \rev{considers} the faithfulness of the reconstructed environment maps, but also explores the ambiguity of the material--light decomposition and the diversity of the solutions.

\begin{figure}
\includegraphics[width=\linewidth]{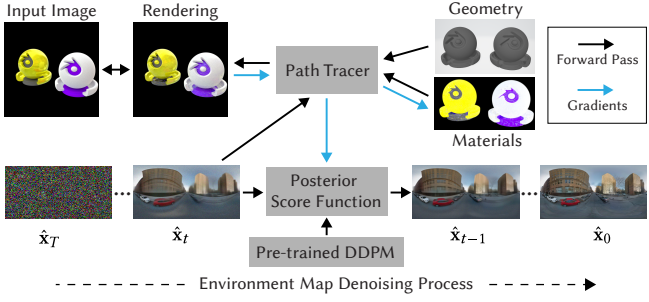} 
\caption
{
Overview of our approach. We first pre-train a DDPM that generates realistic environment maps unconditionally. Then, given input images and geometry, we set up a series of denoising processes. In every time step $t$, our differentiable path tracer takes materials and the posterior estimation of the clean environment map $\hat{x}_t$ as input and produces a rendered image. The gradient from the rendering loss is used to optimize materials and \rev{gets incorporated} into a posterior score function that enforces the DDPM to generate a natural environment map that faithfully explains the input images.
}
\label{fig:pipeline}
\end{figure}

We use a differentiable renderer to guide a pre-trained unconditional diffusion model to sample multiple plausible natural environment maps that can explain the ground-truth observations. At the same time, we optimize the spatially-varying BRDFs of the \rev{scene} for each illumination sample. In the following sections, we first describe the rendering equation, our material model, and our illumination representation (Sec.~\ref{sec:material and light}). Next, we discuss the challenges to extend diffusion posterior sampling (DPS) with a differentiable renderer for inverse rendering problems (Sec.~\ref{sec:dps_inverserendering}), followed by an explanation of the process used to jointly optimize the spatially-varying BRDF and sample plausible environments (Sec.~\ref{sec:joint_opt}).
We give an overview of our pipeline in Fig.~\ref{fig:pipeline}.
%
%
%
\subsection{Rendering Model}
\label{sec:material and light}
We are interested in estimating a natural environment map $\mathbf{x}_0$ and spatially-varying surface materials $\mathbf{k}_{\boldsymbol{arm}}$ explaining input image observations using an analysis-by-synthesis approach.
Global light transport \cite{kajiya1986rendering} under distant illumination can be formulated as
\begin{equation}
\label{eq:rendering equation}
L\left(\mathbf{p}, \boldsymbol{\omega}_{o}\right)=\int_{\Omega_+} \mathbf{x}_0(\boldsymbol{\omega}_{i})\mathbf{T}(\mathbf{p}, \boldsymbol{\omega}_{o},\boldsymbol{\omega}_{i},\mathbf{k}_{\boldsymbol{arm}}) \left(\boldsymbol{\omega}_{i} \cdot \mathbf{n}\right) d \boldsymbol{\omega}_{i}, 
\end{equation}
where $L\left(\mathbf{p}, \boldsymbol{\omega}_{o}\right)$ is radiance leaving surface point $\mathbf{p}$ in direction $\boldsymbol{\omega}_{o}$, $\Omega_+$ is the hemisphere centered at the surface normal $\mathbf{n}$, and 
$\mathbf{T}(\mathbf{p},\boldsymbol{\omega}_{o},\boldsymbol{\omega}_{i},\mathbf{k}_{\boldsymbol{arm}})$
is the radiance transfer function, describing how distant illumination from direction $\boldsymbol{\omega}_{i}$ is scattered through the scene (via potentially many bounces) to finally leave at $\mathbf{p}$ in direction $\boldsymbol{\omega}_{o}$.
We implement $T$ using path tracing \cite{kajiya1986rendering}, which stochastically samples light paths to obtain a Monte Carlo estimate of the global illumination, including potentially complex inter-reflections.

We utilize the state-of-the-art differentiable path tracer Mitsuba3 \cite{jakob2022mitsuba3} with path replay backpropagation \cite{Vicini2021PathReplay}. 
As our research focuses on investigating the ambiguity involved in disentangling material and illumination, we assume that the scene geometry is available.
To represent the material properties, we adopt the principled BSDF in Mitsuba3 \cite{jakob2022mitsuba3} based on the Disney BSDF \cite{burley2012physically,burley2015extending}. 
Our approach utilizes a 256x256 texture $\mathbf{k}_{\boldsymbol{arm}}$ to model RGB albedo $\boldsymbol{a}$, roughness $\boldsymbol{r}$, and metallic properties $\boldsymbol{m}$. 
We use four light bounces in all our experiments, which \rev{we found enough to capture all relevant light transport in the scenes we consider, faithfully modeling global-illumination effects.}

%
%
%
%

%
\subsection{DPS for Inverse Rendering}
\label{sec:dps_inverserendering}
For our application, as the measurement operator $\mathcal{R}(\cdot)$, we use a differentiable path tracer with additional material and camera pose input. 
Namely, the forward measurement process is described as
%
%
\begin{equation}
\label{eq:rendering function}
\mathbf{y}=\mathcal{R}(\mathbf{x}_0,\mathbf{k}_{\boldsymbol{arm}},\boldsymbol{c})+\mathbf{n}_r,\quad\mathbf{y},\mathbf{n}_r\in\mathbb{R}^n,\:\mathbf{x}\in\mathbb{R}^d.
\end{equation}
%
%
This corresponds to the rendering equation (Eq.~\ref{eq:rendering equation}), where $\mathbf{y}$ is the rendered image, $\mathcal{R}(\cdot):\mathbb{R}^d\mapsto\mathbb{R}^n$ is the rendering function, $\boldsymbol{c}$ are camera parameters, and $\mathbf{n}_r$ is noise caused by Monte Carlo sampling.

\rev{Note that the original DPS formulation} requires a known and fixed forward operator $\mathcal{M}(\cdot)$ (Eq.~\ref{eq:dps_measurement}) to approximate the posterior score function $\nabla_{\mathbf{x}_t}\log p_t(\mathbf{x}_t|\mathbf{y})$, which is not applicable in an inverse rendering scenario where the material properties are generally unknown. In our case, the posterior score function is rewritten as 
%
%
\begin{equation}
\label{eq:dps_joint_opt}
\nabla_{\mathbf{x}_t}\log p_t(\mathbf{x}_t|\mathbf{y})\simeq \boldsymbol{s}_{\theta}\left(\mathbf{x}_t,t\right)-\rho\nabla_{\mathbf{x}_t}\|{\hat{\mathbf{y}}_i  - \mathcal{R}(\hat{\mathbf{x}}_t,\mathbf{k}_{\boldsymbol{arm}},\boldsymbol{c}_i)}\|_2^2 \,.
\end{equation} 
%
%
Here, $\hat{\mathbf{x}}_t$ is the posterior estimation of the environment map image from Eq.~\ref{eq:posterior_mean}, \rev{and $\rho$ is a weight hyper-parameter.} Ideally, the sampled environment map $\hat{\mathbf{x}}_t$ should minimize the re-render loss %
%
\begin{equation}
\label{eq:de-render loss}\|{\hat{\mathbf{y}}_i  - \mathcal{R}(\hat{\mathbf{x}}_t,\mathbf{k}_{\boldsymbol{arm}},\boldsymbol{c}_i)}\|_2^2 \,,
\end{equation}
%
and \rev{we opt to concurrently} optimize the material $\mathbf{k}_{\boldsymbol{arm}}$ with the denoising process for the environment map.
To achieve this objective, we propose a novel approach that enables simultaneous sampling of the environment map image and optimization of the material properties in Sec.~\ref{sec:joint_opt}.

DPS provides measurements  (Eq.~\ref{eq:dps_measurement}) that exhibit various types of noise, which correspond to different closed forms of the posterior score function (Eq.~\ref{eq:posterior_score_function}). 
As reported by \citet{lehtinen2018noise2noise}, the use of Monte Carlo sampling in the ray tracer can result in random noise with no specific distribution characteristics. Despite this, we found that applying the posterior score function for Gaussian noise produces satisfactory results for the inverse rendering problem.

%
%
\begin{figure*}[h!]
\includegraphics[width=\textwidth]{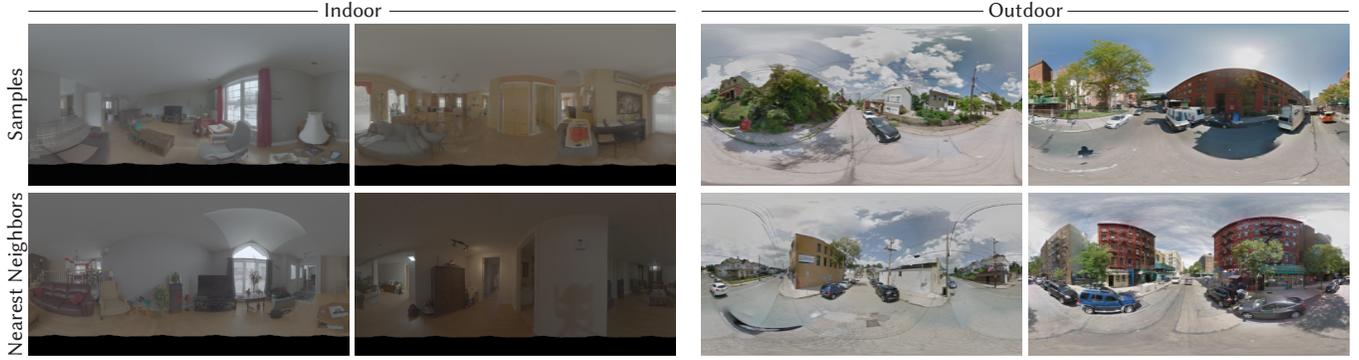} 
\caption
{
Unconditional samples from our generative models (top row) alongside their nearest neighbors in the training dataset (bottom row).
}
\label{fig:dataset}
\end{figure*}
%
%

%
%
\subsection{Joint Optimization}
\label{sec:joint_opt}
Given the posed input image(s) and pre-computed geometry, we first jointly optimize the material and the environment map using Mitsuba3~\cite{jakob2022mitsuba3} with the objective
%
%
\begin{equation}
\label{eq:path tracer loss}
    \mathcal{L}_\text{PT} ({\mathbf{x},\mathbf{k}_{\boldsymbol{arm}}}) = \sum_{i=1}^m  ||{\hat{\mathbf{y}}_i  - \mathcal{R}(\mathbf{x},\mathbf{k}_{\boldsymbol{arm}},\boldsymbol{c}_i)}||^2 \,,
\end{equation}
%
as the initialization, where $\mathbf{x}$ is the reconstructed environment map by Mitsuba3~\cite{jakob2022mitsuba3}, $\hat{\mathbf{y}}_i$ is an input image and $\boldsymbol{c}_i$ is the corresponding camera pose. 
During this initialization stage, we use the Adam \rev{\cite{kingma2014adam}} optimizer with a learning rate of $1e^{-2}$ for metallic and roughness parameters, the environment map, and the base color.

Then we keep the optimized roughness and metallic texture map, but \rev{reset the base color to zero} as our initialization condition for the joint optimization. We find that the initialization of roughness and metallic helps the material optimization converge faster. Next, we gradually denoise the environment map images that fit the observation from our DDPM prior and optimize our estimated material at the same time.
We use a DDPM posterior sampling process (Eq.~\ref{eq:post_sde_reverse}) for 1000 time steps, with the posterior score function in Eq.~\ref{eq:dps_joint_opt}. 

The hyper-parameter $\rho$ in Eq.~\ref{eq:dps_joint_opt} balances the faithfulness of the generation and the quality of reconstruction. A too small $\rho$ encourages the posterior sampling to be as realistic as possible, \rev{disregarding the input observations, while a too large $\rho$ tends to produce results faithful to the observations, yet with inferior realism}.
 In our setting, for the first 500 denoising steps ($t>500$), we set $\rho=0.1$. For time steps $t<500$, $\rho=0.1 \sim 1$, and \rev{in every denoising step, we optimize  the base color, roughness, and metallic jointly with the loss function} 
%
%
\begin{equation}
\label{eq:denoise loss}
    \mathcal{L}_\text{denoise} ({\mathbf{k}_{\boldsymbol{arm}},t}) =\sum_{i=1}^m  ||{\hat{\mathbf{y}}_i  - \mathcal{R}(\hat{\mathbf{x}}_t,\mathbf{k}_{\boldsymbol{arm}},\boldsymbol{c}_i)}||^2 \,.
\end{equation}
%
%

The intuition behind the adjustment of the step size $\rho$  in Eq.~\ref{eq:dps_joint_opt} is that at the early stage of the generation, we encourage the posterior score function to push the sampling distribution towards a more realistic manifold. And when the DDPM adds more details to the images, we raise the step size $\rho$ to force the sampling to fit the observation. At the late stage of the generation where the noise in the sampling is gradually reduced, the estimated environment map from a realistic prior will prevent the optimization for the material from \rev{getting stuck} in a local minimum. 

Finally, after generating a clean and realistic environment map $\mathbf{x}_0$, the material parameters will be refined by 
%
%
\begin{equation}
\label{eq:refine loss}
    \mathcal{L}_\text{refine} ({\mathbf{k}_{\boldsymbol{arm}}}) =\sum_{i=1}^m  ||{\hat{\mathbf{y}}_i  - \mathcal{R}(\mathbf{x}_0,\mathbf{k}_{\boldsymbol{arm}},\boldsymbol{c}_i)}||^2 \,.
\end{equation}
%
%
for 200 \rev{additional} iterations.

%
%
\subsection{Seamless Environment Map Generation}
\label{sec:rotation_equivariant}
We aim to develop a generative model capable of sampling seamlessness spherical environment maps $\mathbf{x}_0$.
This entails constructing a model that effectively captures the distribution of the input data while maintaining seamless panoramas under continuous rotations.

To achieve this, we employ two techniques. Firstly, we apply data augmentation during training. Specifically, when training our DDPM using real-world environment map datasets, we introduce horizontal rotations to each sample, randomizing the rotation angle. By incorporating this data augmentation strategy, our training objective becomes
%
%
\begin{equation}
\label{eq:our_ddpm_loss}
\mathcal{L}(\theta):=\mathbb{E}_{t, \mathbf{x}_0, \boldsymbol{\epsilon},\phi}\left[\left\|\boldsymbol{\epsilon}-\boldsymbol{\epsilon}_\theta\left(\sqrt{\bar{\alpha}_t} \phi(\mathbf{x}_0)+\sqrt{1-\bar{\alpha}_t} \boldsymbol{\epsilon}, t\right)\right\|^2\right],
\end{equation}
%
%
where $\phi$ is the random horizontal rotation operator. This augmentation strategy introduces variations in the rotation angle, effectively reducing potential bias in the input data distribution where the environment maps are predominantly positioned at specific rotation angles. Next, during the denoising generation steps, we employ a similar rotation scheme that replaces the pre-trained unconditional score function (Eq.~\ref{eq:score_function}) as:
%
%
\begin{equation}
\label{eq:rotated_score_function}
\boldsymbol{s}^{'}_{\theta}(\mathbf{x}_t,t)={\phi}_{t}^{-1}\left(\boldsymbol{s}_{\theta}({\phi}_t(\mathbf{x}_t),t)\right),
\end{equation}
%
%
where ${\phi}_t$ is the random horizontal rotation operator at time step t and ${\phi}_{t}^{-1}$ is its reverse. During the denoising process, the score function network $\boldsymbol{s}_{\theta}$ takes the rotated noised images as input and preserves continuity and meaningful information around the stitched seam, resulting in panoramas without noticeable disruptions or artifacts at the boundaries.


\section{Experiments}
\label{sec:results}

Here, we report results of the experiments we conducted to evaluate our method.
We first provide details on our implementation and training procedure (Sec.~\ref{sec:details}).
Then we evaluate the quality of our generative model (Sec.~\ref{sec:sampling}), before providing an in-depth evaluation of the inverse-rendering capabilities of our method (Sec.~\ref{sec:inverse_rendering} and Sec.~\ref{sec:qualitative}).
We \rev{provide further analyses (Sec.~\ref{sec:analysis}) and} conclude with a short section on additional applications (Sec.~\ref{sec:applications}).


\subsection{Implementation and Training Details}
\label{sec:details}

Since indoor and outdoor illumination conditions exhibit markedly different characteristics, we employ two separate DDPM models.
For the indoor model, we train on the Laval indoor dataset \cite{gardner2017learning}, which includes 2.2k high dynamic range (HDR) environment maps, while for the outdoor model, we use the Streetlearn dataset \cite{mirowski2019streetlearn}, containing 143k low dynamic range (LDR) environment maps.
\rev{ We apply an HDR reconstruction pre-process on the outdoor dataset using the method of \citet{Marcel:2020:LDRHDR}.}
We perform data augmentation by applying random horizontal shifts (corresponding to horizontal rotations of the scene-enclosing sphere) and horizontal flips.
For stable training, we compress the HDR content using the global invertible mapping
$f(\mathbf{x}) = 0.5 \mathbf{x}^{\frac{1}{2.4}}$ for outdoor data and  $0.9 \mathbf{x}^{\frac{1}{6}}$ for indoor.
Before using our samples for path tracing, we apply $f^{-1}$ to recover HDR illumination.
We use the network architecture proposed by \citet{nichol2021improved} without modifications.
Training one model takes 1 week using four A100 GPUs.
Obtaining a single sample from our joint optimization procedure takes \rev{20 minutes, including 1000 iterations of interleaved environment map denoising (290\,ms) and material optimization (630\,ms)}. We provide all source code and pre-trained models on \url{https://vcai.mpi-inf.mpg.de/projects/2023-DPE/}.



\subsection{Quality of the Generative Models}
\label{sec:sampling}

%
%
\begin{table}[h!]
	\caption{
     \rev{Quality (FID$\downarrow$) of generative models for unconditional sampling.}
     }
    \centering
    \begin{tabular}{lrrr}
          Model & \texttt{GAN} & \texttt{RENI} &\textbf{Ours} \\
          \toprule
          Indoor & 45.5 &288.2  & \textbf{12.3}  \\
          Outdoor & 15.8 &323.7  & \textbf{7.6}  \\
          \bottomrule
    \end{tabular}
    \label{table:sample_quality}
\end{table}
%
%

Before we investigate the inverse-rendering capabilities of our method, we seek to gain insights into the quality of our trained generative models.
To this end, we consider unconditional sampling.
Tab.~\ref{table:sample_quality} provides a numerical evaluation using FID~\cite{heusel2017gans} scores.
In Fig.~\ref{fig:dataset} we show samples of our models.
To make sure that our networks do not simply memorize the training data, we also display the respective nearest neighbors in the training dataset. 
Distances between samples are computed using the LPIPS \cite{zhang2018perceptual} metric, while also taking into account the augmentations (rotations, flips) performed during training.
We see that our samples are markedly different from the closest examples in the training data corpus.
We conclude that our models produce high-quality samples, which look realistic and capture the data distribution well.

%
%
%
\begin{figure*}[!htp]
\includegraphics[width=\linewidth]{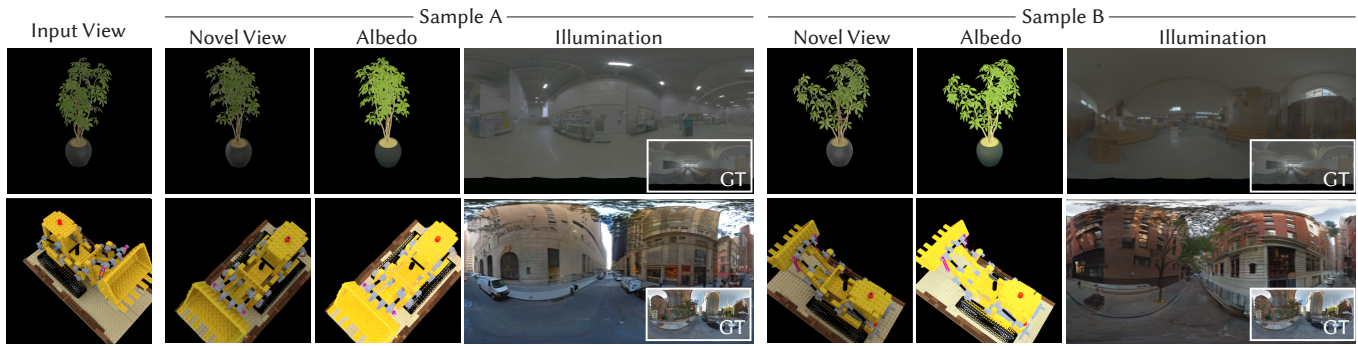} 
\caption
{
Qualitative results on synthetic datasets.
Here we show two samples per scene, demonstrating indoor (top) and outdoor (bottom) illumination.
Notice that the recovered illumination samples are markedly dissimilar in the details, but share the overall illumination structure, faithfully capturing the ambiguity of the inverse problem.
Insets show the ground-truth environment maps.
}
\label{fig:synthetic_samples}
\end{figure*}
%
%
%


\subsection{Inverse Rendering}
\label{sec:inverse_rendering}

We now turn to our core task: The ambiguity-aware decomposition of a scene into illumination and material properties, based on one or multiple images.
We assume we have access to or can reconstruct the scene geometry before running our method.
This step is orthogonal to our approach, and we give details on how geometry is obtained for each experimental setup.


\subsubsection{Quantitative Evaluation}

We start by considering a set of synthetic scenes, where ground-truth geometry, illumination, and materials are available for evaluation.
Specifically, we base our analysis on nine scenes, including six scenes from the NeRF \cite{mildenhall2021nerf} dataset with ten illumination conditions per scene (five indoor and five outdoor), all distinct from the training data corpus.
The scenes exhibit a broad spectrum of material types.
Inputs to our method are $16\sim50$ multi-view images of resolution 800x600 pixels.


\paragraph{Metrics}
To quantitatively evaluate the obtained solutions, we are interested in several properties, listed as columns in Tab.~\ref{table:main_quant_indoor} and Tab.~\ref{table:main_quant_outdoor}:
First, we consider three different aspects concerning the reconstructed \emph{materials}.
We measure albedo reconstruction accuracy using the mean squared error (MSE).
We also show how much albedo reconstruction varies when running the method multiple times with exactly the same input by computing diversity in the form of variance $\sigma_\text{sample}^2$ across five runs.
Further, we evaluate how far the reconstruction of albedo is invariant under different illumination conditions, computed as the variance $\sigma_\text{invar}^2$.

Second, we analyze the reconstructed \emph{environment maps}.
We start with the common full-reference quality metrics PSNR, SSIM \cite{wang2004image}, and LPIPS \cite{zhang2018perceptual} to gain an understanding of how close the reconstructions are to the ground-truth illumination.
We are further interested in measuring the naturalness of the obtained illumination conditions.
To this end, we employ the FID \cite{heusel2017gans} score, comparing the conditional distribution of reconstructed environment maps given the input images to the unconditional distribution of environment maps in the training data corpus.
The distributions are clearly dissimilar by construction, and the number of samples is low for our reconstructions, naturally resulting in relatively high scores, but we found the FID to nevertheless correlate well with the perceived naturalness of the solutions.
As an additional indicator of naturalness we compute the non-reference HDR metric PU21-NIQE \cite{mittal_niqe,hanji2022comparison}.
Analogous to the albedo analysis, we consider the diversity of the illumination estimates by computing variance $\sigma_\text{sample}^2$ across different samples.

Finally, we show full-reference quality metrics for the \emph{novel-view synthesis} task on held-out input views.
This helps us understand how accurately our decompositions explain the scene's appearance.

\rev{
Notice that the true values of $\sigma_\text{sample}^2$ (both for material and illumination) and $\sigma_\text{invar}^2$ should reflect the variance of the true posterior distribution, which is unknown.
A reasonably low value for $\sigma_\text{invar}^2$ indicates stability, a reasonably high value for $\sigma_\text{sample}^2$ indicates diversity.
However, blindly aiming for a minimal $\sigma_\text{invar}^2$ or maximal $\sigma_\text{sample}^2$ is not desirable.
As a counter-example, consider a deterministic method that always converges to the same albedo estimation and therefore has $\sigma_\text{invar}^2 = 0$, which is clearly a sub-optimal solution.
}

%
%
\begin{figure*}[!htb]
\includegraphics[width=\linewidth]{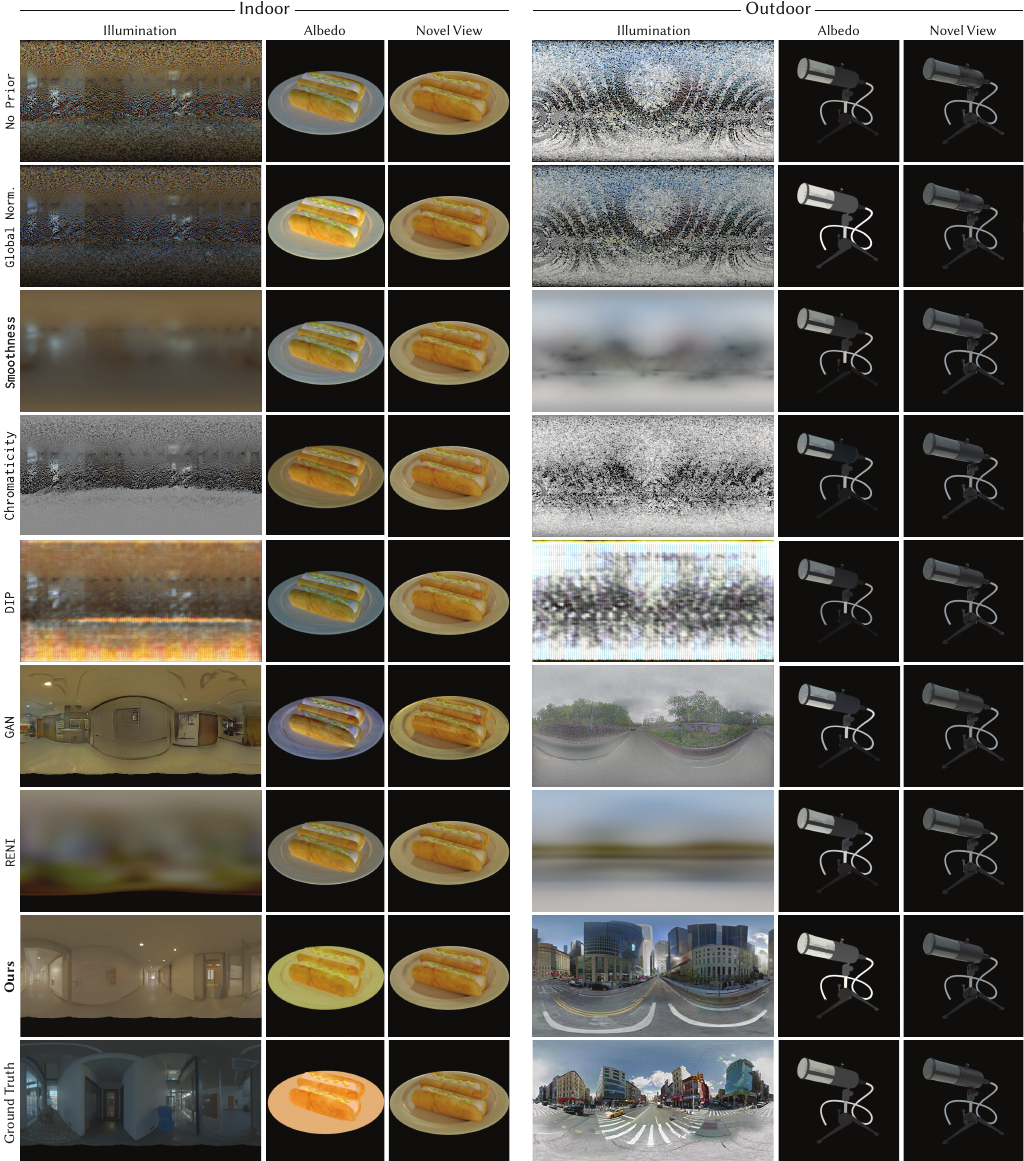} 
\caption
{
Scene decompositions using different priors for two synthetic scenes with indoor (left) and outdoor (right) illumination.
}
\label{fig:results_priors}
\end{figure*}
%
%

\paragraph{Baselines}
We compare our diffusion-based prior on illumination to a broad variety of alternatives, including data-agnostic and data-driven priors.
Whenever necessary, we set weighting factors balancing reconstruction and regularization such that both losses have the same magnitude.
We consider the following priors:

%
%
\begin{table*}[htb!]
	\centering
    \caption{Quantitative evaluation of \textbf{indoor} illumination.
    \rev{For each prior/method, we present statistics on reconstructed \emph{Albedo}, \emph{Illumination}, and novel \emph{View Synthesis}.
    Our method outperforms all baselines by estimating more accurate albedos and illumination, and consequently synthesizing more accurate novel views.
    The measures $\sigma_\text{invar}^2$ and $\sigma_\text{sample}^2$ give an indication of stability and diversity, respectively, but their ground-truth values are unknown.
    }
    }
    \scalebox{1}{
    \begin{tabular}{lrrrrrrrrrrrr}
        \multirow{2}{*}{Prior/Method} & \multicolumn{3}{c}{Albedo} & \multicolumn{6}{c}{Illumination} & \multicolumn{3}{c}{View Synthesis} \\
        \cmidrule(lr){2-4} \cmidrule(lr){5-10} \cmidrule(lr){11-13}

        & MSE$\downarrow$ & $\sigma_\text{sample}^2$ & $\sigma_\text{invar}^2$ & PSNR$\uparrow$ & SSIM$\uparrow$ & LPIPS$\downarrow$ & FID$\downarrow$ & NIQE$\downarrow$ & $\sigma_\text{sample}^2$ & PSNR$\uparrow$ & SSIM$\uparrow$ & LPIPS$\downarrow$ \\
        
        \toprule

        \texttt{No Prior} & 0.086 & 0.0007 & 0.030 & 16.3 & 0.14 & 0.89 & 329.6 &10.5 &0.0017 & 30.5 &0.947 & 0.027 \\
        \texttt{Global Norm.} &0.078  &0.0006  &0.025  &20.8  &0.38   &0.64   &304.3 &9.4  &0.0001  &30.5   &0.947  &0.027 \\
        \texttt{Smoothness} & 0.081 & 0.0006 & 0.028 & 18.8 & 0.61 & 0.46 & 341.3 &9.26 &0.0001 & 33.8 & 0.964 & 0.024\\
        \texttt{Chromaticity} & 0.085 & 0.0002 & 0.029 & 18.2 & 0.41 & 0.63 & 344.9 &10.5  &0.0001 &32.0 & 0.948 & 0.025 \\
        \texttt{DIP} & 0.123 & 0.0057& 0.038 &11.0 & 0.26 & 0.56 & 301.8 &43.1  & 0.0052 &32.3 &0.943 & 0.029 \\
        \texttt{GAN} & 0.106 & 0.0069 & 0.055 &16.9 & 0.59 & 0.35 & 187.1 &6.98 & 0.0072 & 29.4 & 0.944 & 0.029
        \\
        
        \texttt{RENI} & 0.091 & 0.0215 & 0.043 & 15.9 &0.49 & 0.54 & 371.8 &17.0 & 0.0225 & 27.8 & 0.913 & 0.037 \\
        \midrule
        NeRFactor &0.084  &0.0011  &0.032   &7.3   &0.31  &0.69   &453.7 &18.9  &0.0028   &26.4   &0.918  &0.051 \\
        
        NvDiffRec  &0.075  &0.0001  &0.038  &8.5  &0.17   &0.65   &468.3  &10.6 &0.0049   &29.3   &0.940  &0.049 \\
        \quad w/ GT geom.  &0.063  &0.0001  &0.030  & 10.2  &0.20   &0.61   &415.6 &6.84  &0.0077   &32.7  &0.952  &0.037 \\
        
        NvDiffRecMC   &0.103  &0.0001  &0.076  &14.8  &0.27   &0.55   &322.2  &9.98 &0.0002   &28.1   &0.934  &0.057 \\
        \quad w/ GT geom.  &0.082  &0.0001  &0.049  & 15.4  &0.34   &0.62   &307.8 &6.61   &0.0004   &29.0   &0.939  &0.046 \\
        
         \midrule
        \textbf{Ours} & \textbf{0.033} & 0.0082 & 0.021 & \textbf{21.9} & \textbf{0.67} & \textbf{0.26} & \textbf{135.6} &\textbf{5.14} & 0.0096 & \textbf{34.0} & \textbf{0.967} & \textbf{0.022} \\
        
        \bottomrule
    \end{tabular}
    }
\label{table:main_quant_indoor}
\end{table*}
%
%
\begin{table*}[htb!]
	\centering
    \caption{Quantitative evaluation of \textbf{outdoor} illumination. \rev{Refer to the caption of Tab.~\ref{table:main_quant_indoor} for more details.}}
     \scalebox{1}{
    \begin{tabular}{lrrrrrrrrrrrr}
        \multirow{2}{*}{Prior/Method} & \multicolumn{3}{c}{Albedo} & \multicolumn{6}{c}{Illumination} & \multicolumn{3}{c}{View Synthesis} \\
        \cmidrule(lr){2-4} \cmidrule(lr){5-10} \cmidrule(lr){11-13}

         & MSE$\downarrow$ & $\sigma_\text{sample}^2$ & $\sigma_\text{invar}^2$ & PSNR$\uparrow$ & SSIM$\uparrow$ & LPIPS$\downarrow$ & FID$\downarrow$ & NIQE$\downarrow$ & $\sigma_\text{sample}^2$ & PSNR$\uparrow$ & SSIM$\uparrow$ & LPIPS$\downarrow$ \\
         
        \toprule

        \texttt{No Prior} &0.087  &0.0010  &0.031  &13.0   &0.12   &0.87   &379.6 &8.88  &0.0030   &31.6   &0.939  &0.029  \\
        \texttt{Global Norm.} &0.080  &0.0007  &0.027  &15.1  &0.16   &0.83   &380.5 &8.63  &0.0005    &31.6   &0.939  &0.029 \\
        \texttt{Smoothness} &0.084  &0.0005  &0.031  &14.9   &0.49   &0.64   & 376.3 &7.25 &0.0001   &33.1   &0.960  &0.025\\
        \texttt{Chromaticity} &0.088 &0.0001  &0.032  &13.5    &0.15  &0.81   &398.7 &8.83  & 0.0001  &31.5    & 0.931 &0.029\\
        \texttt{DIP} &0.142  &0.0014  &0.054  &9.7   &0.19   &0.63    &371.5 &106.24  &0.0071  &32.1   &0.927  &0.037 \\
        
        \texttt{GAN} &0.078  &0.0051  &0.094   &15.4   &0.41  &0.50   &163.3 &6.26  &0.0086   &29.7   &0.936  &0.031 \\

        \texttt{RENI} &0.108  &0.0136  &0.090  &11.3   &0.36   &0.76   &487.9 &27.8  &0.0232  & 30.6  &0.931  &0.044 \\

         \midrule
        NeRFactor &0.103  &0.0009  &0.035   &6.4   &0.28  &0.86   &477.6 &21.2  &0.0024   &25.7   &0.912  &0.056 \\
        NvDiffRec  &0.088  &0.0001 &0.037  &8.4  &0.09   &0.70   &448.2  &9.67  &0.0055   &31.0   &0.938  &0.039 \\
        \quad w/ GT geom.  &0.049  &0.0002   &0.032  &9.5  &0.07   &0.63   &392.2 & 6.42 &0.0069   & 32.3  &0.950  &0.037 \\
        
        NvDiffRecMC   &0.185  &0.0001  &0.082  &11.1  &0.13   &0.68   &390.5  & 8.73 &0.0002   &29.8   &0.926  &0.039 \\
        \quad w/ GT geom.  &0.109  &0.0001  &0.044  &13.6  &0.22   &0.69   &350.4 &6.10  &0.0002   &30.1   &0.941  &0.038 \\
        
         \midrule
        \textbf{Ours} &\textbf{0.031}  &0.0069  &0.016 &\textbf{20.0}  &\textbf{0.58}   &\textbf{0.30}   &\textbf{109.1}  &\textbf{4.36} & 0.0108  &\textbf{34.3}   &\textbf{0.968}  &\textbf{0.022} \\
        
        \bottomrule
    \end{tabular}
    }
\label{table:main_quant_outdoor}
\end{table*}
%
%

\begin{itemize}
    \item \texttt{No Prior}: 
    As a baseline, we run path tracing-based inverse rendering without any prior at all.

    \item \texttt{Global Norm.}: 
    To address the inherent \emph{global} ambiguities of illumination--reflectance decomposition, we proceed as in the previous case, but additionally scale the estimated environment maps to enforce their median to match the median of our dataset.
    
    \item \texttt{Smoothness}: 
    A very common prior used in many inverse problems is the assumption that obtained solutions should be smooth \cite{lyu2022neural,zhang2021nerfactor}. We adopt the 2D total variation penalty
    \begin{equation*}
        \mathcal{L}_\text{smooth} =
        \sum_{\mathbf{u}}
        \left\| \nabla_{\mathbf{u}} \mathbf{x}_0(\mathbf{u}) \right\|_1 ,
    \end{equation*}
    where $\mathbf{u}$ denotes the directional coordinates in the environment map.

    \item \texttt{Chromaticity}: 
    Observing that illumination is frequently not very colorful, we follow Munkberg et~al.~\shortcite{Munkberg_2022_CVPR} and use
    \begin{equation*}
        \mathcal{L}_\text{chroma} =
        \left\|
        \sum_{c=1}^{3}
        \left(
        \mathbf{x}_0^c
        -
        \frac{1}{3}\sum_{c=1}^{3}\mathbf{x}_0^c
        \right)
        \right\|,
    \end{equation*}
    where $\mathbf{x}_0^c$ denotes the $c$'s color channel of $\mathbf{x}_0$.

    \item \texttt{DIP}:
    The Deep Image Prior \cite{ulyanov2018deep} represents a signal by optimizing the weights of a CNN with random inputs. This enforces a structural prior, empirically favoring solutions that have natural-image statistics.
    
    \item \texttt{GAN}:
    We train a StyleGAN2-ADA \cite{karras2020training} model on our datasets and optimize over the model's latent space using \rev{Pivotal Tuning Inversion \cite{roich2022pivotal}}.
    As for our method, we train two separate models for indoor and outdoor scenes.

    \item \texttt{RENI}:
    The recent method of Gardner et~al.~\shortcite{gardner2022rotation} uses a rotation-equivariant neural field to train a generative model of environment maps. We follow their method to optimize the latent code of this model.
    Also for this prior, we train two separate models.

\end{itemize}

\rev{The quality of unconditional samples of the \texttt{GAN} and \texttt{RENI} models is evaluated in Tab.~\ref{table:sample_quality}.}

In addition to the priors discussed above, we also compare our method to \rev{three} state-of-the-art full-blown inverse-rendering systems.
First, we consider NvDiffRec \cite{Munkberg_2022_CVPR} and NvDiffRecMC \cite{hasselgren2022shape}.
In addition to the recovery of illumination and reflectance, both methods also optimize for geometry.
For a fair differential comparison, we consider two variants of their methods:
The full version, and replacing their geometry estimation with fixed ground-truth geometry. 
\rev{In addition, we include a comparison with NeRFactor \cite{zhang2021nerfactor} as a representative of inverse rendering using a volumetric representation.}


\paragraph{Results}

\begin{figure}
\includegraphics[width=\linewidth]{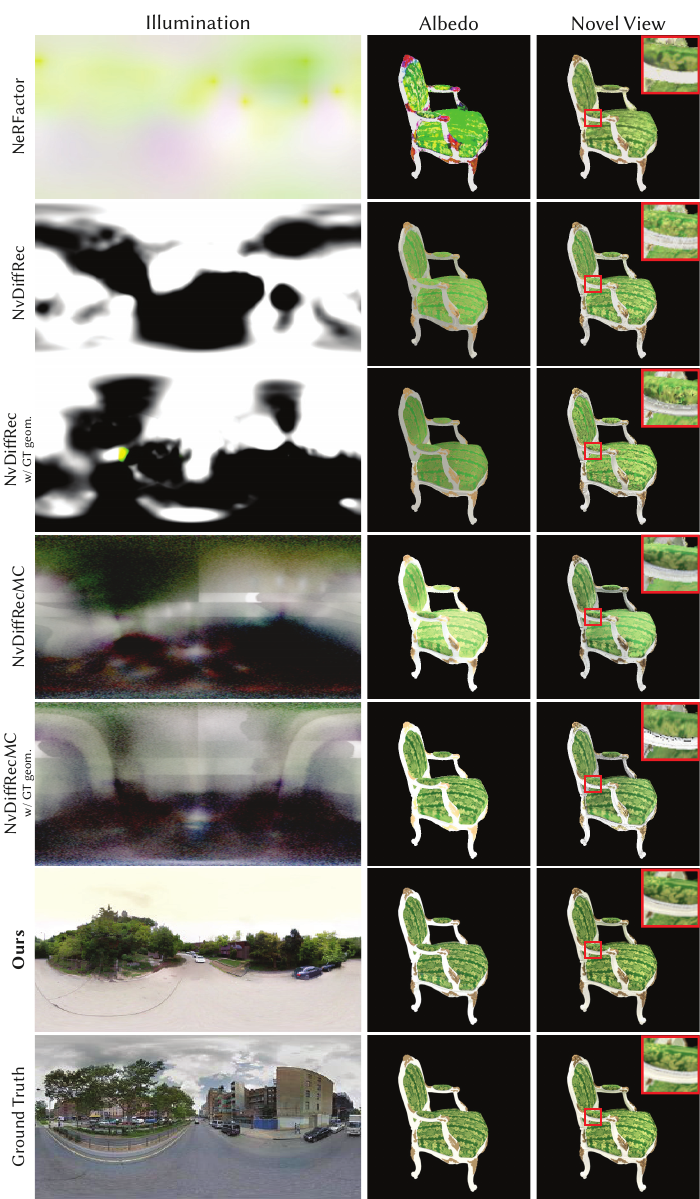} 
\caption
{
\rev{Comparison with baseline differentiable rendering methods.}
}
\label{fig:results_nvidia}
\end{figure}

We list numerical results of our experiments in Tab.~\ref{table:main_quant_indoor} for indoor and Tab.~\ref{table:main_quant_outdoor} for outdoor illumination conditions.
Fig.~\ref{fig:synthetic_samples} displays samples obtained from our method.
A qualitative comparison against the baseline priors in our setup is provided in Fig.~\ref{fig:results_priors}, while Fig.~\ref{fig:results_nvidia} illustrates comparisons against the inverse-rendering systems.
Our supplemental video shows additional qualitative results and comparisons.
\rev{We provide runtime evaluations in the supplemental document.}

\begin{figure}
\includegraphics[width=\linewidth]{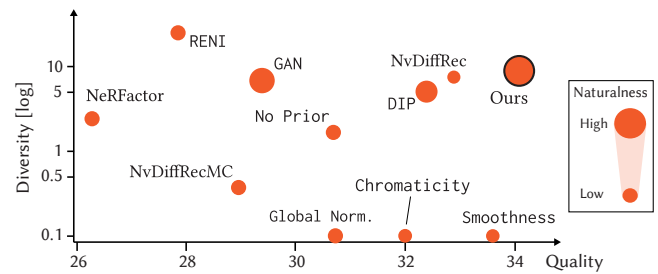} 
\caption
{
Overview of different methods in terms of novel-view synthesis quality (PSNR; x-axis), environment map diversity ($\sigma_\text{sample}^2$; y-axis, log-scale), and environment map naturalness (FID; point size).
Only our solution achieves high scores across all three aspects.
}
\label{fig:quality_vs_diversity}
\end{figure}

We observe that our method excels in reconstructing environment maps and materials, where the latter tend to be \rev{reasonably} stable across different illumination conditions.
More importantly, though, our environment maps are significantly more \emph{natural} than all baselines and achieve high \emph{diversity} scores, while at the same time resulting in the highest novel-view synthesis \emph{quality}.
Fig.~\ref{fig:quality_vs_diversity} illustrates the distribution of methods according to these three core criteria.
While the \texttt{GAN} prior achieves competitive scores in terms of diversity and naturalness, it fails to properly explain the input data.
On the other end of the spectrum, the \texttt{Smoothness} prior achieves high-quality novel-view synthesis results (explaining its popularity in the inverse-rendering community), but always converges to the same solution while producing rather unnatural results.
NvDiffRec achieves competitive results in terms of novel-view synthesis quality and diversity, but also suffers from rather unrealistic environment maps.

%
%
%
%

%
%
%
%

\subsection{Qualitative Evaluation}
\label{sec:qualitative}

%
%
\begin{figure*}
\includegraphics[width=\linewidth]{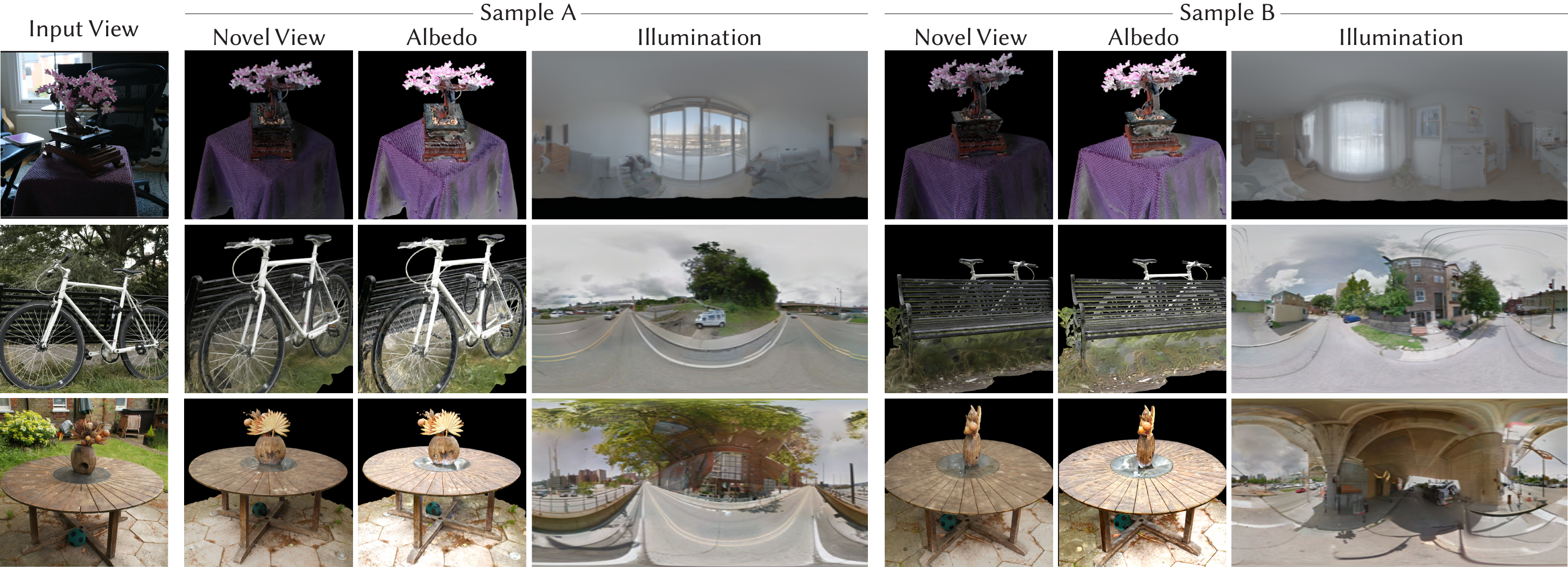} 
\caption
{
Qualitative results on real-world datasets.
Here we show two samples per scene.
Notice that our optimization does only have access to the foreground object, but still manages to synthesize environment maps that match the overall color distribution of the background, while exhibiting a high level of realism.
}
\label{fig:real_world_results}
\end{figure*}
%
%

To evaluate our method on real scenes, we consider a set of scenes from the MipNeRF-360 dataset \cite{barron2022mipnerf360}, \rev{using 100 images per scene.}
We first run NeuS \cite{wang2021neus} on the multi-view images to obtain geometry of the foreground objects as well as a foreground--background separation, \rev{and automatically generate the texture map for the mesh extracted from the SDF from Blender’s ``Smart UV Project'' operator \cite{Blender}}.
We then run our optimization, while masking out the background in the input images.
Fig.~\ref{fig:real_world_results} shows the results of this experiment, which illustrate that we can sample high-quality decompositions.
The supplemental \rev{document and} video provide more results.



\subsection{Analysis}
\label{sec:analysis}

Here, we investigate several aspects of our method in more detail.

\begin{figure}
\includegraphics[width=\linewidth]{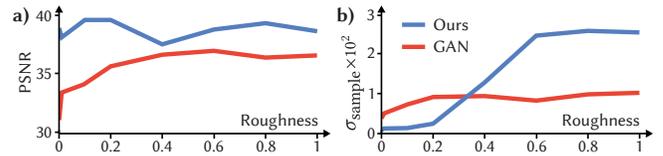} 
\caption
{
\rev{View synthesis quality (a) and environment map diversity (b)} as a function of material roughness.
Our approach consistently outperforms the \texttt{GAN} prior in terms of quality and exhibits diversity only for ambiguous scenes.
}
\label{fig:roughness_vs_variance}
\end{figure}

\begin{figure}
\includegraphics[width=\linewidth]{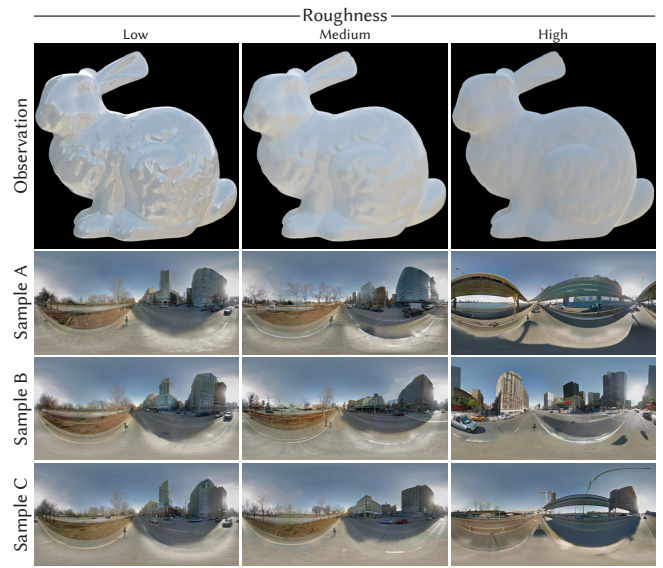} 
\caption
{
\rev{Qualitative results with three levels of material roughness. We show three illumination samples from our method for each roughness level. While all samples are very similar for the highly specular material, their diversity increases as the material approaches diffuse appearance.}
}
\label{fig:diversity_bunny}
\end{figure}

First, we consider illumination reconstruction quality and diversity as a function of material roughness.
Intuitively, as the roughness of scene materials is increased towards diffuse appearance, we expect an increasing amount of ambiguity during illumination reconstruction, as the environment map is convolved with a progressively larger BRDF lobe.
To systematically investigate the behavior of our method in this regard, we set up a synthetic scene containing a Stanford Dragon to be rendered from three views.
We then run a sequence of experiments, where we perform a sweep over the roughness parameter of the dragon's material, with five illumination conditions per roughness setting.
In Fig.~\ref{fig:roughness_vs_variance} we investigate the behavior of our method and compare it against the \texttt{GAN} prior.
In Fig.~\ref{fig:roughness_vs_variance}a we observe a consistently high-quality reconstruction result across all roughness levels, while the \texttt{GAN} prior exhibits overall \rev{inferior} results, struggling in particular with highly specular materials.
Fig.~\ref{fig:roughness_vs_variance}b reveals that, \rev{in contrast to the \texttt{GAN} prior,} the diversity of our samples strongly correlates with roughness, satisfying our expectations.
For example, diversity is almost zero when the material is purely specular, which is expected behavior since the input images essentially contain a distorted reflection of the environment map, eliminating uncertainty.
\rev{In Fig.~\ref{fig:diversity_bunny} we provide a corresponding qualitative study.}

\begin{figure}
\includegraphics[width=\linewidth]{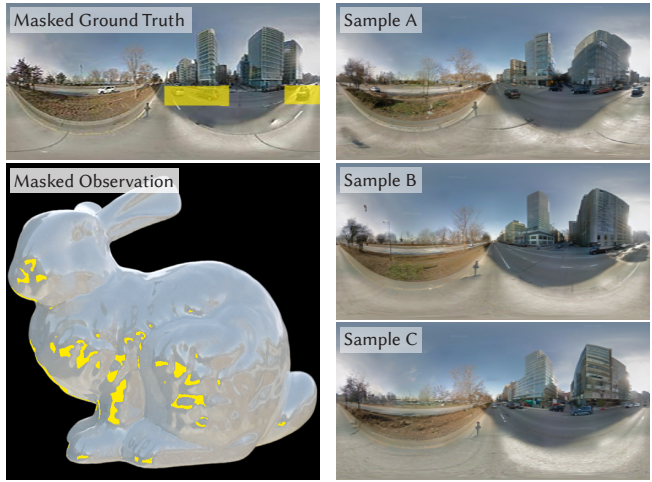} 
\caption
{
\rev{Environment map sampling from masked observations on specular materials. 
We mask out cars in the ground-truth environment map (top left) and mark all corresponding pixels in the observations (bottom left).
During training, these pixels are ignored.
Our samples (right) match the observable areas while generating distinct details in the unseen parts.}
}
\label{fig:masking_bunny}
\end{figure}

\rev{
In Fig.~\ref{fig:masking_bunny} we study the behavior of our method in the presence of highly specular materials:
We mask out regions in the ground-truth environment map (top left in  Fig.~\ref{fig:masking_bunny}) and record which pixels in the image observations are affected by this mask (bottom left in Fig.~\ref{fig:masking_bunny}).
We then run our method on the observations, but exclude the masked-out pixels from receiving path-tracer gradients.
We see that the samples from our method reconstruct the ground-truth environment map quite faithfully while hallucinating plausible content in the masked regions (right in Fig.~\ref{fig:masking_bunny}).}
We further visualize the amount of information that can be extracted from highly specular materials in the supplemental document.

%
%
%
%

%
%
%
%

%
%
%
%

\begin{figure}
\includegraphics[width=\linewidth]{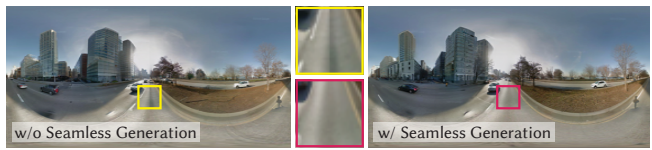} 
\caption
{
Ablation: Seamless environment map generation.
}
\label{fig:ablation_seamless}
\end{figure}

We perform an ablation analysis of our seamless environment map generation process (Sec.~\ref{sec:rotation_equivariant}) in Fig.~\ref{fig:ablation_seamless}.
We see that our method is effective in removing seams in the samples.

%
%

%
%

\begin{figure}
\includegraphics[width=\linewidth]{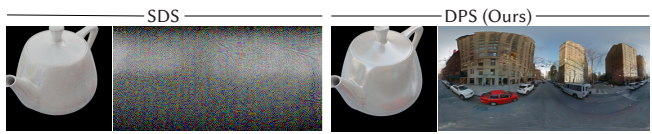}
\vspace{-0.6cm}
\caption
{
Ablation: Training with a score distillation sampling (SDS) loss instead of our DPS-based formulation.
For each configuration, we show a reconstructed view (left) alongside an environment map sample (right).
}
\label{fig:sds_ablation}
\vspace{-0.4cm}
\end{figure}

We are further interested in validating the choice of DPS as our solver.
In Fig.~\ref{fig:sds_ablation} we compare our approach against a best-effort result based on score distillation sampling (SDS) \cite{poole2022dreamfusion}.
In each time step, we add a loss term that moves the current state of the optimized environment map closer to higher-density regions learned by the pre-trained diffusion model. 
We first add a random amount of noise to the current estimate of the environment map. 
The loss function then computes the difference between the estimated and ground truth noise. 
As proposed by \citet{poole2022dreamfusion}, we do not backpropagate through the diffusion model itself. 
%
We observe that the SDS results are consistently inferior to the ones from our DPS-based solution, failing to produce realistic samples.
%

%
%

%
%


\subsection{Applications}
\label{sec:applications}

Our scene decompositions can be used for a variety of applications.
In Fig.~\ref{fig:teaser} we show an example of material editing and relighting.
In this case, geometry was reconstructed using multiple views, but input to our system is a single view of the almost diffuse object, resulting in substantial illumination ambiguity.
We see that changing the original material to a translucent one produces a plausible appearance, while the \texttt{No Prior} or \texttt{Smoothness} solutions result in noise and loss of overall structure, respectively.

\section{Discussion and Conclusion}

We have presented a novel approach that addresses the ambiguity problem in inverse rendering.
At the core of our method is a denoising diffusion probabilistic model trained on natural environment maps.
On the technical level, we have extended diffusion posterior sampling towards an optimizable measurement operator, which allows for the simultaneous reconstruction of illumination and materials.
We have shown that our recovered environment maps have high quality and high diversity while exhibiting an unprecedented level of naturalness.

Our approach is not free from limitations, providing plenty opportunity for future work.
We did not aim at devising a full inverse-rendering system, as we assume geometry to be given.
This has allowed us to study the problem of material--illumination ambiguities in depth.
A natural next step is the inclusion of geometry estimation into our pipeline.

\begin{figure}
\includegraphics[width=\linewidth]{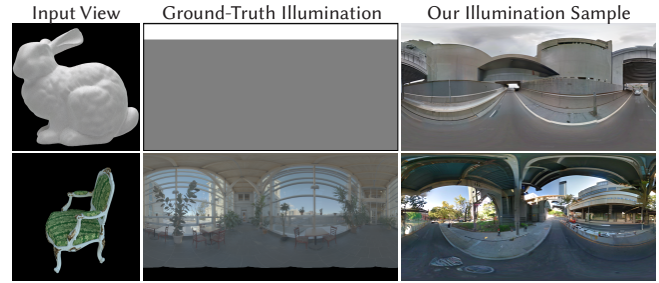} 
\vspace{-0.6cm}
\caption
{
Limitation: If the illumination of the scene does not follow the data distribution used to train our diffusion model, our samples can be far from the ground truth, while still explaining the input observations.
}
\label{fig:limitation}
\vspace{-0.4cm}
\end{figure}

Importantly, we can only expect reasonable results if the distribution of environment maps captured by our generative model matches the task.
Fig.~\ref{fig:limitation} illustrates an example of an intentionally created mismatch.
In the top row, we use a simple synthetic environment map to illuminate a bunny and run our method using the outdoor model.
In the bottom row, we illuminate an armchair using indoor illumination and, again, run our method using the outdoor model.
We see that, in these cases, the recovered environment maps are far from the ground truth, but nevertheless capture the overall illumination condition well.

Current diffusion models are notoriously inefficient when it comes to generating samples.
We inherit this limitation, which makes our (unoptimized) implementation roughly 5x slower than a vanilla path-tracing-based solution.

We hope to inspire future work on generative inverse rendering, balancing the intricate triangle between accuracy, naturalness, and ambiguity.

%
%

%
%


\textbf{Acknowledgements.}

This work was supported by the ERC Consolidator Grant 4DReply (770784) and the Lise Meitner Postdoctoral Fellowship.


\bibliographystyle{ACM-Reference-Format}
\bibliography{ms}
\end{document}


\vspace{4ex}
\begin{flushleft}\LARGE
\textbf{Diffusion Posterior Illumination for Ambiguity-aware Inverse Rendering -- Supplemental Material}
\end{flushleft}


\section{Performance Breakdown }
\label{sec:performance_breakdown}
 We report runtime performance comparisons in Tab.~\ref{table:run_time}.
%
%
\begin{table}[h!]
	\begin{center}
	\centering
    
   \begin{tabular}{lrrr}
			%
			%
            \toprule
   
			\textbf{Prior}                                             & Time (ms) / iteration & Iterations & Total time  (min)  	\\ 

			\midrule
			\texttt{No Prior} &669 & 400  & 4.5    \\
            \texttt{Global Norm.} &669 & 400  & 4.5   \\
            \texttt{Smoothness} &675  &400   &4.5 \\
            \texttt{Chromaticity} &675  &400  &4.5\\
            \texttt{DIP} &698 &600  &7.0 \\

            \texttt{GAN} &706  &950  &11.2  \\

            \texttt{RENI} &690 &1000  &11.5 \\                            
		  \texttt{Ours} &920 & 1000 &15.3 (19.8) 	\\     
			%
			%

        \bottomrule
		\end{tabular}
  
          \end{center}
  
  \caption{Runtime performance comparisons. We report the denoising and joint optimization scheme as \texttt{Ours}. Our method includes  4.5 minutes for initialization and material refinement, and 15.3 minutes for optimization, resulting in 19.8 minutes in total.
            }  
\label{table:run_time}
\end{table}
%
%
\section{Extractable Information from Highly Specular Materials}
\label{sec:diff_bunny}

\begin{figure*}[h!]
\includegraphics[width=\linewidth]{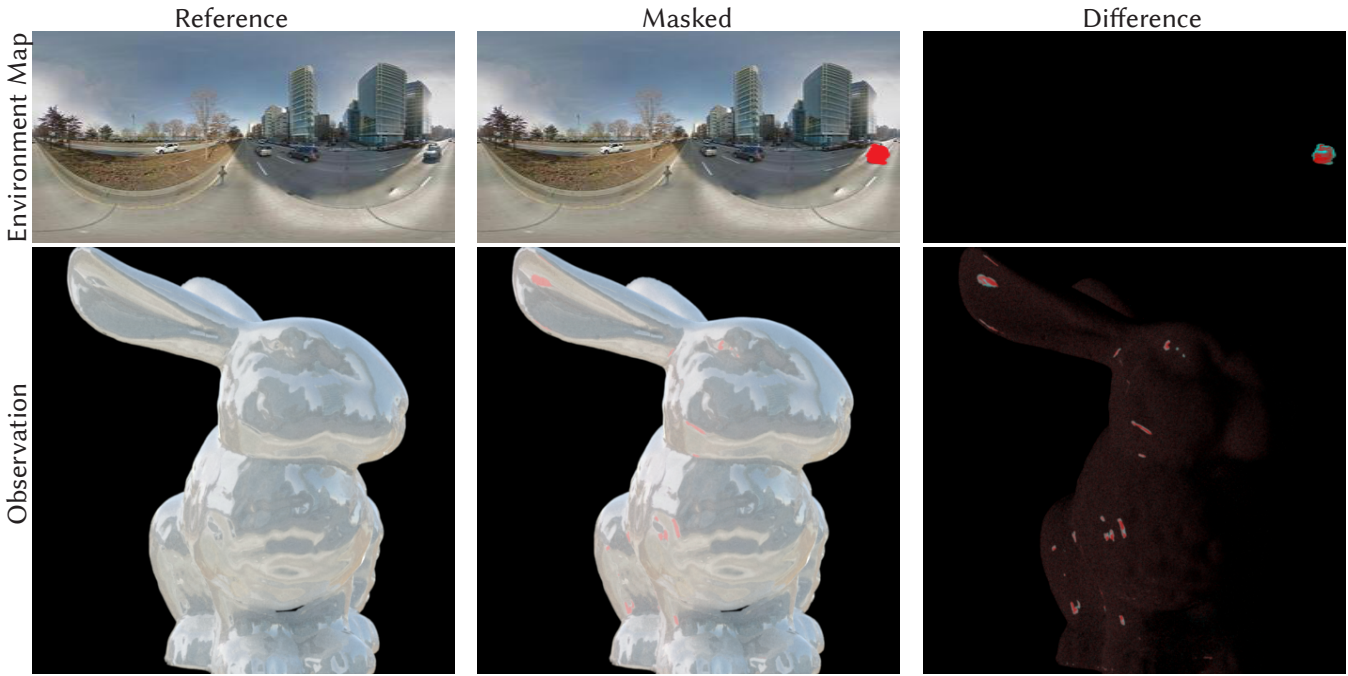} 
\caption
{
\textcolor{black}{Masking pixels in an environment map reveals that there is significant information to reconstruct close-to-ground-truth illumination from observations with highly specular materials.}
}
\label{fig:diff_bunny}
\end{figure*}

\textcolor{black}
{
%
In Fig.~\ref{fig:diff_bunny}, we analyze the information that can be extracted from highly specular materials, where small changes in lighting lead to significant changes in the image observations.
In the figure, replacing a car in the environment map with a pure red box leads to many differences in the rendered images and shows that the renderings contain significant information about the car.
Our method can use this information to reconstruct environment maps close to ground truth. 
}

%
%
%
%

\section{Evaluation on Real-world Datasets}
\label{sec:real_world_comparison}

\begin{figure*}[bp!]
\includegraphics[width=\linewidth]{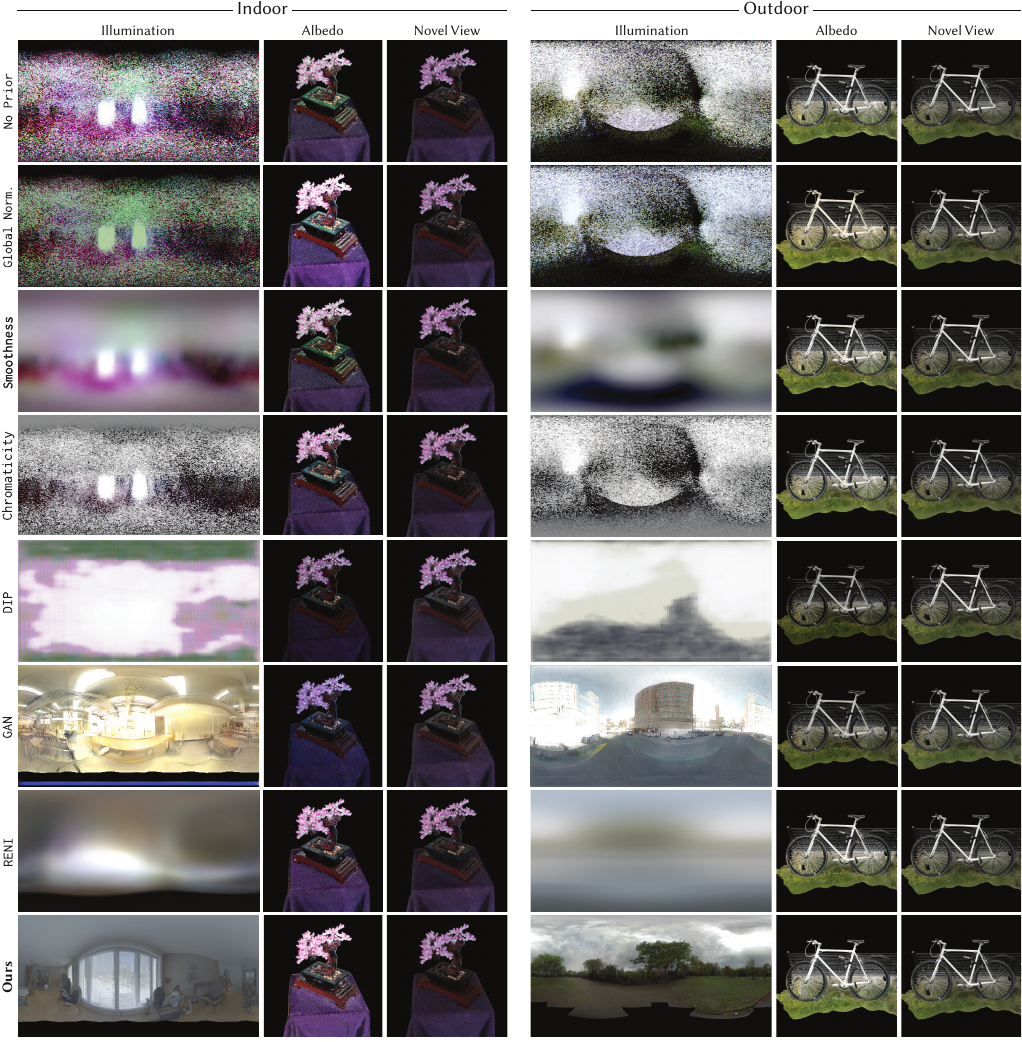} 
\caption
{
Scene decompositions using different priors for two real-world scenes with indoor (left) and outdoor (right) illumination.
}
\label{fig:results_realworld_comparison}
\end{figure*}

In Fig.~\ref{fig:results_realworld_comparison} we provide qualitative comparisons against baseline priors on two real-world datasets.

%
%
%
%